\documentclass[10pt,letterpaper]{article}
\usepackage[top=0.85in,left=2.75in,footskip=0.75in]{geometry}

\usepackage{amsmath,amssymb}

\usepackage{changepage}

\usepackage{textcomp,marvosym}
\usepackage[utf8]{inputenc}
\usepackage[T1]{fontenc}
\usepackage{float}
\usepackage{algorithm}
\usepackage{algpseudocode}
\usepackage{float}
\usepackage{tabularx}
\usepackage{booktabs}
\usepackage{cite}

\usepackage{nameref,hyperref}

\usepackage[right]{lineno}

\usepackage[nopatch=eqnum]{microtype}
\DisableLigatures[f]{encoding = *, family = * }

\usepackage[table]{xcolor}

\usepackage{array}

\newcolumntype{+}{!{\vrule width 2pt}}

\newlength\savedwidth

\raggedright
\usepackage[aboveskip=1pt,labelfont=bf,labelsep=period,justification=raggedright,singlelinecheck=off]{caption}

\makeatletter
\renewcommand{\@biblabel}[1]{\quad#1.}
\makeatother

\usepackage{lastpage,fancyhdr,graphicx}
\usepackage{epstopdf}
\fancyheadoffset[L]{2.25in}
\fancyfootoffset[L]{2.25in}
\begin{document}
\vspace*{0.2in}

\begin{flushleft}
{\Large
\textbf\newline{Early Mortality Prediction in ICU Patients with Hypertensive Kidney Disease Using Interpretable Machine Learning} 
}
\newline
\\
Yong Si\textsuperscript{1},
Junyi Fan\textsuperscript{1},
Li Sun\textsuperscript{1},
Shuheng Chen\textsuperscript{1},
Minoo Ahmadi\textsuperscript{1},
Elham Pishgar\textsuperscript{2},
Kamiar Alaei\textsuperscript{3},
Greg Placencia\textsuperscript{4},
Maryam Pishgar\textsuperscript{1*}

\bigskip
\textbf{1} Department of Industrial and Systems Engineering, University of Southern California, 3715 McClintock Ave GER 240, Los Angeles, 90087, California, United States
\\
\textbf{2} Colorectal Research Center, Iran University of Medical Sciences, Tehran Hemat Highway next to Milad Tower, Tehran, 14535, Iran
\\
\textbf{3} Department of Health Science, California State University, Long Beach (CSULB), 1250 Bellflower Blvd, Long Beach, 90840, California, United States
\\
\textbf{4} Department of Industrial and Manufacturing Engineering, California State Polytechnic University, Pomona, 3801 W Temple Ave, Pomona, 91768, California, United States
\\
\bigskip

%
%





* pishgar@usc.edu

\end{flushleft}
\section*{Abstract}
\textbf{Background:} Hypertensive kidney disease (HKD) represents a clinically vulnerable subgroup in intensive care units (ICUs), yet risk stratification tools tailored to this population are lacking. Timely identification of patients at high risk of short-term mortality is essential for informed clinical decision-making and resource allocation.

\textbf{Methods:} We developed a machine learning-based framework to predict 30-day in-hospital mortality in ICU patients with HKD using early clinical data from the MIMIC-IV v2.2 database. A cohort of 1366 adult patients was curated by applying strict inclusion criteria and excluding confounding malignancy cases. Eighteen features—spanning vital signs, laboratory results, comorbidities, and therapeutic interventions—were selected through a two-step process combining random forest importance and mutual information filtering. Multiple models were trained and compared using stratified five-fold cross-validation, with CatBoost achieving the highest performance.

\textbf{Results:} The CatBoost model achieved an AUROC of 0.88 on the independent test set, with sensitivity of 0.811 and specificity of 0.798, ensuring high recall for critical cases. Interpretability tools such as SHAP values and Accumulated Local Effects (ALE) plots confirmed the model's reliance on physiologically meaningful predictors, including altered consciousness, vasopressor requirement, and coagulation status. Furthermore, we integrated the DREAM algorithm to estimate patient-specific posterior risk distributions, allowing clinicians to assess not only the predicted mortality but also its associated uncertainty.

\textbf{Conclusions:} This study presents a clinically grounded and interpretable machine learning pipeline that enables early, real-time risk assessment for ICU patients with HKD. By incorporating both high predictive performance and uncertainty quantification, the model supports individualized triage and facilitates transparent, evidence-based decision-making at the bedside. This approach has strong potential for clinical deployment and warrants external validation in diverse critical care settings.



\section*{Introduction}

Hypertension, commonly known as high blood pressure, is a chronic medical condition characterized by persistently elevated arterial blood pressure, which is defined as a systolic blood pressure $\geq$ 140 mmHg and a diastolic blood pressure $\geq$ 90 mmHg.\cite{world2023global,williams20182018,carey2018prevention} Moreover, the pathophysiology of hypertension is complex and multifactorial, involving a combination of genetic predispositions and environmental factors that contribute to dysregulation of the Renin-Angiotensin-Aldosterone System (RAAS), endothelial dysfunction, increased sympathetic nervous system activity, and structural changes in the vasculature\cite{carey2018prevention,harrison2013mosaic,harrison2021pathophysiology,martyniak2022new}. Consequently, hypertension affects an estimated 1.28 billion adults aged 30-79 years globally, making it one of the most prevalent non-communicable diseases worldwide\cite{zhou2021worldwide,wierzejska2020global}. Furthermore, the economic burden associated with hypertension is substantial, encompassing direct healthcare costs related to diagnosis, treatment, and management, as well as indirect costs stemming from lost productivity due to morbidity and premature mortality\cite{kearney2005global,forouzanfar2017global,egan2019global,mills2020global}. 

If left unmanaged, hypertension can lead to severe complications, including myocardial infarction, stroke, heart failure, kidney disease, and peripheral artery disease. Among all these complications, chronic kidney disease is particularly noteworthy due to its bidirectional relationship with hypertension. Hypertension is a leading cause of kidney disease; for instance, it is the second leading cause of kidney failure in the United States after diabetes, and its prevalence is significantly higher in hypertensive individuals, with approximately 1 in 5 adults (20\%) in the US with high blood pressure also having kidney disease \cite{cheung2019blood, centers2019chronic,lai2025global}. Hypertension contributes to glomerulosclerosis, tubulointerstitial fibrosis, and progressive decline in renal function. Conversely, as kidney function deteriorates, the ability to regulate blood pressure is severely compromised, often leading to resistant hypertension. In patients with kidney disease, the prevalence of apparent treatment-resistant hypertension can be as high as 39\% to 55\% depending on the definition and kidney disease stage \cite{an2022prevalence,bansal2024revisiting,yu2020bidirectional}. This elevated blood pressure further accelerates kidney damage, contributing to a vicious cycle between the two conditions and significantly increasing cardiovascular risk \cite{ravarotto2018oxidative, clementi2021holistic}. This intricate interplay thus highlights a critical need for advanced strategies in the integrated management of Hypertensive Kidney Disease (HKD), particularly within the intensive care unit (ICU) setting, to effectively guide patient prognosis, facilitate early monitoring, and ultimately improve therapeutic outcomes and optimize healthcare resource allocation.

Xia et al. (2022) proposed a Cox regression-based nomogram model to predict the 1-year and 3-year mortality risk in patients with hypertensive chronic kidney disease (HKD) admitted to ICU\cite{xia2022survival}. Their methodology involved identifying predictive factors using univariate and stepwise Cox regression analyses on a cohort from the MIMIC-III database, followed by internal and external validation of the nomogram. Achieving an area under the receiver operating characteristic (ROC) curve of 0.775 for 1-year mortality and 0.769 for 3-year mortality in external validation, their model identified insurance status, albumin levels, alkaline phosphatase, mean corpuscular hemoglobin concentration, mean corpuscular volume, history of coronary angiogram, hyperlipemia, digoxin medication, acute renal failure, and history of renal surgery as top predictors.

Despite the significant burden of hypertension and its well-established role in the development and progression of kidney disease, studies specifically investigating mortality predictors in critically ill patients with confirmed HKD, particularly within the ICU setting, remain relatively sparse. While numerous studies have explored mortality prediction in broader cohorts of Chronic Kidney Disease (CKD) or acute kidney injury (AKI) patients \cite{hansrivijit2021prediction,chen2025machine}, or in general hypertensive populations \cite{fan2025predicting}, the unique pathophysiological nuances and clinical trajectory of HKD patients in the highly acute ICU environment may necessitate a more targeted predictive approach. Existing research, such as that by Xia et al. (2022)\cite{xia2022survival}, has initiated efforts to predict outcomes in this specific cohort, demonstrating the feasibility of identifying relevant prognostic factors. However, there is a clear opportunity to further explore and refine predictive models utilizing advanced methods to provide more precise and actionable insights for this high-risk subgroup.

Recent years have witnessed machine learning (ML) models making remarkable strides in predicting clinical outcomes, offering sophisticated approaches to leverage complex medical data for enhanced prognostic insights. Among these, CatBoost, a gradient boosting decision tree algorithm, stands out for several compelling advantages in the clinical prediction domain: (1) it efficiently handles categorical features automatically, reducing preprocessing needs and overfitting through "Ordered Target Encoding"; (2) it offers superior predictive accuracy and robustness to overfitting due to its "Ordered Boosting" technique; and (3) it provides built-in feature importance calculations, enhancing model interpretability, which is crucial for clinical adoption. Recent studies applying CatBoost for various clinical outcome predictions further validate its efficacy \cite{si2025retrospective,chen2025interpretable,ashrafi2025enhanced}. These aforementioned characteristics and demonstrated applications collectively position CatBoost as a highly attractive candidate for developing precise and actionable predictive models for complex conditions like HKD in the ICU, where data quality, robustness, and interpretability are paramount for improving treatment strategies and optimizing healthcare resource allocation.

This study introduces several key innovations that distinguish our approach to early-phase mortality risk prediction in ICU patients with Hypertensive Kidney Disease HKD:
\begin{itemize}
    \item \textbf{Clinical and Medical Significance}: This is the first study to develop a tailored risk stratification tool for short-term (14-day) mortality specifically in the clinically vulnerable subgroup of HKD patients within the intensive care unit (ICU) setting. By utilizing routinely available clinical data within 24 hours of ICU admission, our work directly supports timely identification of high-risk patients for informed clinical decision-making and individualized interventions.

    \item \textbf{Pipeline Advantages}: This study developed an interpretable CatBoost-based model leveraging a comprehensive set of routinely collected clinical data, including demographics, vital signs, laboratory results, medications, and comorbidities, all within the critical first 24 hours of ICU admission. This approach ensures applicability and relevance in real-world critical care scenarios.

    \item \textbf{Model Performance}: Our CatBoost model achieved superior predictive performance for 14-day mortality with an impressive AUROC of 0.880 (95\% CI: 0.838–0.919). This was rigorously validated by outperforming several established baseline models, including XGBoost, LightGBM, Random Forest, Logistic Regression, and Lasso Regression, demonstrating its robust and effective prognostic capability.

    \item \textbf{Interpretability Insights}: Employing advanced explainability methods, specifically feature importance analysis and SHapley Additive exPlanations (SHAP) values, this study identified critical predictors of early ICU mortality in HKD patients. Key predictors included age, Sequential Organ Failure Assessment (SOFA) score, Glasgow Coma Scale (GCS), and vasopressor use, providing actionable insights that can guide clinical management strategies for this complex patient population.
\end{itemize}

\section*{Materials and Methods}
\subsection*{Data Source and Study Design}

This retrospective cohort study was conducted using the Medical Information Mart for Intensive Care IV (MIMIC-IV, version 2.2), a publicly available, de-identified database developed by the Massachusetts Institute of Technology and the Beth Israel Deaconess Medical Center. MIMIC-IV contains detailed electronic health records from over 60,000 ICU admissions between 2008 and 2019, encompassing demographics, vital signs, laboratory tests, medications, procedures, and outcomes. All analyses adhered to the MIMIC-IV data use agreement, and no IRB approval was required due to prior de-identification.

The study aimed to predict 30-day in-hospital mortality in ICU patients with hypertensive kidney disease using early clinical data. A multi-stage pipeline was implemented, including patient selection, data preprocessing, feature engineering, model training, statistical evaluation, and interpretability analysis.

From an initial cohort of 35,794 ICU admissions, only the first ICU stay per patient was included to avoid duplication bias and ensure independence of observations. Adult patients ($\geq$18 years) diagnosed with hypertensive kidney disease (HKD) were identified using standardized ICD-9 and ICD-10 codes. To minimize confounding from cancer-related prognostic factors, individuals with active malignancies or metastatic tumors were excluded. All candidate predictors—including demographics, comorbidities, vital signs, laboratory values, procedures, and medication administrations—were extracted from the first 24 hours of ICU admission to capture early-phase clinical information. The outcome of interest was 30-day in-hospital mortality, determined using discharge disposition and recorded time to death.

Preprocessing involved median imputation for continuous variables and mode imputation for categorical ones. Dynamic physiologic features were summarized using minimum, maximum, and mean values. All continuous features were rescaled using min-max normalization. SMOTE was applied within training folds to address class imbalance, and preprocessing steps were performed only on training data to avoid leakage.

Initial feature selection started with over 400 candidates across six clinical categories. A two-stage selection procedure was applied: random forest importance scores were used to rank features based on Gini impurity, followed by mutual information filtering to remove weakly dependent variables. Eighteen final features were retained based on statistical relevance and clinical validation.

Six models—CatBoost, LightGBM, XGBoost, logistic regression, Naïve Bayes, and shallow neural networks—were trained with five-fold cross-validation and grid search for hyperparameter tuning. Model evaluation focused on AUROC, with secondary metrics including sensitivity, specificity, F1-score, and predictive values. To enhance interpretability and clinical utility, the study incorporated ablation analysis, SHAP, ALE plots, and posterior uncertainty estimation using the DREAM algorithm.

The complete modeling pipeline is summarized in Algorithm~\ref{alg:HKD_mortality}, designed to ensure methodological rigor, clinical relevance, and real-world applicability in ICU decision-making for HKD patients.

\begin{algorithm}[H]
\caption{\textbf{Machine Learning Pipeline for 30-Day Mortality Prediction in HKD ICU Patients}}
\label{alg:HKD_mortality}
\begin{algorithmic}[1]
\Require MIMIC-IV (v2.2) ICU data from 2008--2019
\Ensure Prediction of 30-day in-hospital mortality

\State \textbf{Step 1: Cohort Construction}
\State Identify hypertensive kidney disease (HKD) using ICD-9/10 codes
\State Retain first ICU admission per patient
\State Exclude: age < 18, malignancy/metastatic cancer

\State \textbf{Step 2: Data Extraction and Preprocessing}
\State Extract variables from first 24h: chartevents, labevents, procedures, drugs, admissions
\State Impute numeric missing values with median
\State Impute categorical missing values with mode
\State Derive min, max, mean for dynamic physiologic variables
\State Normalize continuous variables using min-max scaling

\State \textbf{Step 3: Feature Selection}
\State Begin with 400+ features across six clinical domains
\State Apply Random Forest to compute Gini importance
\State Remove features with near-zero importance
\State Compute Mutual Information scores between features and mortality label
\State Remove features with negligible MI values
\State Retain 18 clinically validated final features

\State \textbf{Step 4: Class Imbalance Handling}
\State Apply SMOTE to oversample mortality cases within training folds

\State \textbf{Step 5: Model Training and Evaluation}
\State Split data into training (70\%) and testing (30\%) sets
\ForAll{model $\in$ \{CatBoost, LightGBM, XGBoost, Logistic Regression, Naïve Bayes, Shallow NN\}}
    \State Perform stratified 5-fold cross-validation
    \State Conduct grid search for hyperparameter optimization
    \State Evaluate performance: AUROC, F1-score, sensitivity, specificity, PPV, NPV
\EndFor

\State \textbf{Step 6: Statistical and Interpretability Analysis}
\State Perform t-tests on key features (train vs. test)
\State Conduct ablation study to assess feature contribution to AUROC
\State Visualize non-linear effects via Accumulated Local Effects
\State Interpret individual predictions using SHAP
\State Estimate uncertainty with posterior sampling`

\State \textbf{Step 7: Final Reporting}
\State Calculate AUROC confidence intervals via 2000 bootstrap replicates
\State Summarize findings for clinical deployment and decision support
\end{algorithmic}
\end{algorithm}

\subsection*{Patient Selection}

This study employed a methodologically rigorous, multi-stage patient selection framework designed to construct a clinically homogeneous cohort for developing machine learning models to predict 30-day in-hospital mortality among ICU patients with hypertensive kidney disease (HKD). Our selection strategy was guided by clinical specificity, methodological rigor, and generalizability to real-world critical care settings.

We systematically queried the MIMIC-IV database (version 2.2), identifying 35,794 unique ICU encounters from Beth Israel Deaconess Medical Center between 2008 and 2019. To ensure analytical independence and eliminate bias from repeated measurements, we retained only the first ICU admission for each unique patient, reflecting the clinical reality that initial presentations capture the most acute and treatment-naïve physiological state.

From this deduplicated cohort, we employed standardized ICD-9 and ICD-10 diagnostic codes to identify 1,809 patients with documented hypertensive kidney disease, representing 5.1\% of all ICU admissions. The clinical rationale for focusing exclusively on HKD patients stems from their unique pathophysiological complexity, where sustained hypertension leads to progressive nephrosclerosis and impaired autoregulation. In the ICU context, these patients exhibit heightened vulnerability to acute decompensation due to compromised physiological reserve and increased susceptibility to multi-organ dysfunction.

The final exclusion criterion involved systematic removal of patients with active malignancies or metastatic cancer (n = 175), yielding our definitive analytical cohort of 1,366 patients. This exclusion represents 12.8\% of the age-eligible HKD population and reflects a critical methodological decision to enhance model specificity. Cancer patients in the ICU present fundamentally different mortality determinants, including treatment-related toxicities, palliative care considerations, and disease-specific prognostic factors that may overwhelm traditional physiological predictors.

\begin{figure}[H]
\centering
\includegraphics[width=0.65\linewidth]{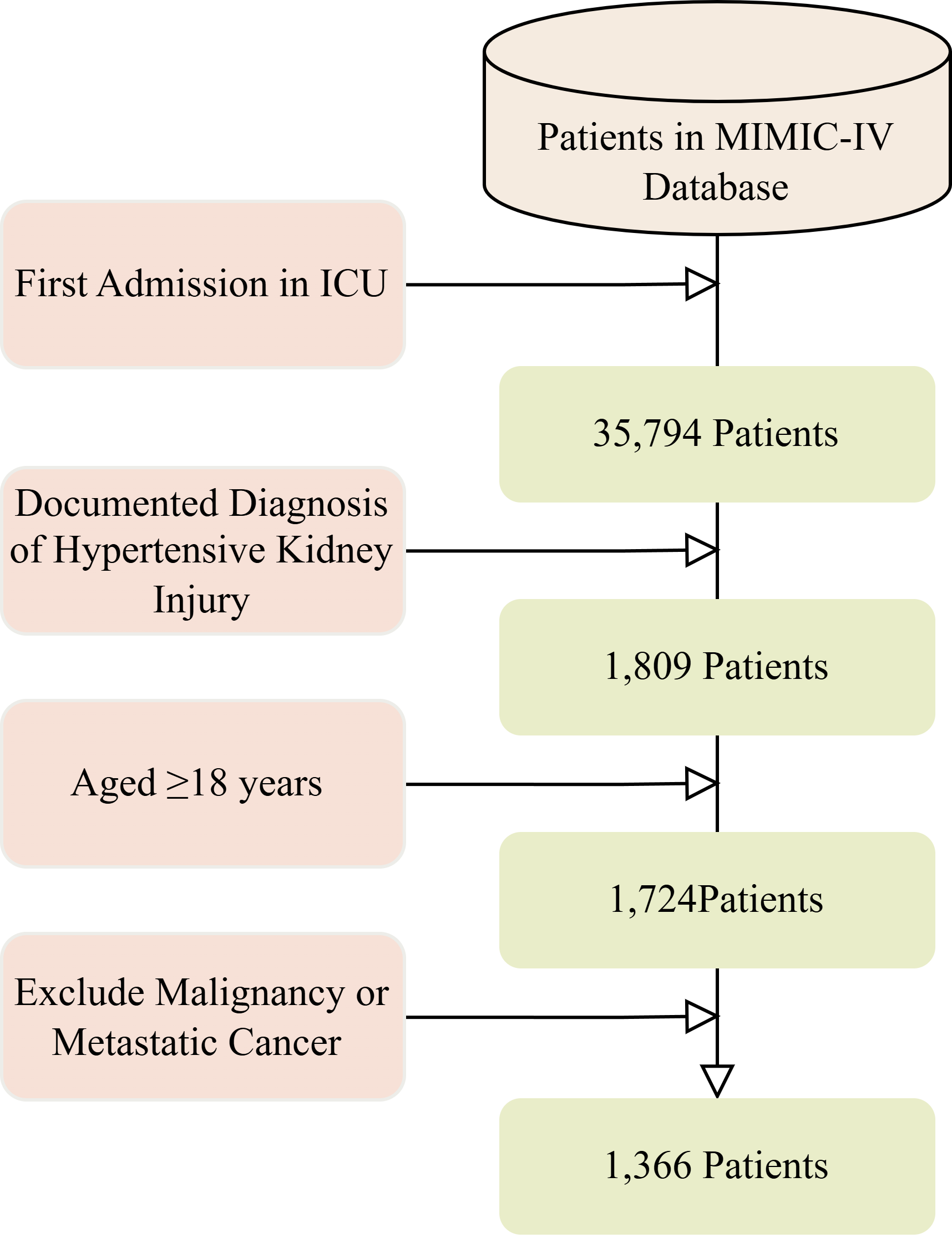}
\caption{\textbf{Patient selection process for hypertensive kidney injury cohort}}
\label{fig:patient_selection}
\end{figure}

\subsection*{Data Preprocessing}

In this study, we developed a systematic preprocessing pipeline specifically tailored to the complexities of real-world ICU data. Critical care datasets present unique challenges including incomplete documentation due to clinical urgency, heterogeneous monitoring frequencies across different physiological systems, and the inherent class imbalance where adverse outcomes like mortality represent a minority of cases. Our preprocessing approach addresses these clinical realities while maintaining the integrity of the predictive model.

Missing data in the ICU setting often reflects clinical decision-making rather than random omission. For example, certain laboratory tests may not be ordered for stable patients, while frequent vital sign monitoring may be interrupted during procedures. We addressed this clinical reality using evidence-based imputation strategies.

For laboratory values and continuous physiological measurements such as \textit{phosphate, anion gap, lactate, PTT}, and \textit{respiratory rate}, we applied median imputation. This approach is particularly suited to ICU data because it remains robust to the extreme values commonly encountered in critical care (e.g., severely elevated lactate in shock states or very low blood pressures during hemodynamic instability). The median represents a clinically reasonable "typical" value that does not distort the physiological range.

For categorical clinical assessments such as \textit{Richmond-RAS scale} (sedation level) and \textit{GCS motor response} (neurological function), we used mode imputation to preserve the most frequently observed clinical state. This reflects standard ICU practice where clinical assessments often default to expected ranges when specific documentation is unavailable.

Critical care patients exhibit rapid physiological changes that single-point measurements may miss. To capture the dynamic nature of ICU monitoring, we extracted temporal summaries (minimum, maximum, and mean values) for key time-varying parameters during the first 24 hours of admission. This approach mirrors clinical practice where intensivists assess both acute derangements (minimum/maximum values) and overall trends (mean values).

For instance, \textit{heart rate alarm-low events} indicate episodes of bradycardia that may signal impending cardiac arrest, while \textit{respiratory rate variability} can indicate respiratory distress or ventilator weaning readiness. The mathematical representation captures these clinical patterns:

\begin{equation}
x_{i,\text{mean}} = \frac{1}{T} \sum_{t=1}^{T} x_{i}^{(t)}, \quad 
x_{i,\text{max}} = \max_{t \in [1,T]} x_{i}^{(t)}, \quad 
x_{i,\text{min}} = \min_{t \in [1,T]} x_{i}^{(t)}
\end{equation}

where $T$ represents the monitoring period and $x_{i}^{(t)}$ represents the measured value at time $t$.

Laboratory values and physiological parameters span vastly different scales—from Glasgow Coma Scale scores (3-15) to serum creatinine levels (0.5-20+ mg/dL) to APACHE scores (0-71). Without standardization, variables with larger numerical ranges would disproportionately influence the model, potentially masking clinically important but numerically smaller parameters.

We applied min-max normalization to rescale continuous variables to a [0,1] range, ensuring that each clinical parameter contributes proportionally to the prediction regardless of its measurement scale. This included laboratory values like \textit{lactate} (critical for sepsis assessment), severity scores like \textit{APACHE III}, and temporal factors like \textit{ED duration} (which may indicate diagnostic complexity):

\begin{equation}
x_{\text{normalized}} = \frac{x - x_{\text{min}}}{x_{\text{max}} - x_{\text{min}}}
\end{equation}

Binary clinical indicators require no transformation as they directly represent clinical decision points. Procedural interventions such as \textit{multi-lumen catheter placement} indicate central venous access for complex monitoring or therapy, while medication administration like \textit{norepinephrine} signals vasopressor-dependent shock. These 0/1 indicators preserve their clinical meaning while providing clear input to the predictive model.

Critical comorbidities including \textit{cerebral edema} and \textit{severe sepsis with septic shock} were similarly encoded as binary features. These conditions significantly alter mortality risk and represent distinct pathophysiological processes that intensivists routinely consider when assessing prognosis.

ICU mortality, while clinically significant, occurs in a minority of admissions, creating a statistical challenge known as class imbalance. Traditional machine learning models may develop a bias toward predicting survival, potentially missing patients at highest risk for death.

To address this clinical reality, we employed the Synthetic Minority Over-sampling Technique (SMOTE) within our cross-validation framework. SMOTE generates synthetic examples of high-risk patients by interpolating between existing mortality cases, helping the model learn to identify subtle patterns associated with poor outcomes:

\begin{equation}
x_{\text{synthetic}} = x_{\text{mortality case}} + \delta \cdot (x_{\text{similar case}} - x_{\text{mortality case}})
\end{equation}

where $\delta \in [0, 1]$ is a random coefficient. Importantly, this augmentation was applied only to training data, ensuring that our performance evaluation reflects real-world conditions rather than artificially enhanced scenarios.

All preprocessing steps were designed to preserve clinical interpretability while enhancing predictive accuracy. The pipeline was applied exclusively to training data, with preprocessing parameters (such as median values for imputation) derived from the training set and then applied to validation data. This approach simulates real-world deployment where the model must make predictions on new patients without access to future outcomes.

This preprocessing framework ensures that our mortality prediction model reflects the complex, dynamic nature of ICU care while maintaining the statistical rigor necessary for reliable clinical decision support. The resulting dataset provides a robust foundation for identifying patients at risk for adverse outcomes in the critical care setting.

\subsection*{Feature Selection}

We began with over 400 candidate features derived from the first 24 hours of ICU stay, compiled based on clinical literature and expert consultations. To ensure interpretability and facilitate structured modeling, these features were categorized into six clinically meaningful groups: (1) vital signs from chartevents, including blood pressure, respiratory rate, and neurologic scores, were incorporated to reflect the patient’s immediate physiological status, which is essential for identifying early signs of hemodynamic compromise;\cite{ripolles2023hypotension} (2) laboratory measurements from labevents, such as lactate, bicarbonate, and coagulation parameters, provided objective markers of renal function, acid–base balance, and systemic derangements commonly seen in hypertensive kidney injury;\cite{chen2019acid} (3) medications and interventions extracted from procedureevents and prescriptions, including norepinephrine administration and multi-lumen catheter insertion, served as indirect indicators of illness severity and clinical escalation; \cite{bellomo2008vasoactive}(4) comorbidities derived from diagnosis records, including cerebral edema and septic shock, were included to capture background disease burdens that compound the risk of mortality in critically ill patients with HKD;\cite{zhang2021trends} (5) administrative features such as emergency department (ED) duration served as proxies for care delays or triage inefficiencies prior to ICU admission\cite{siddiqi2023clinical}; and (6) severity scores like apsiii were incorporated as composite metrics of overall acuity with well-validated predictive power in critical care settings\cite{griffin2017hypertensive}. This clinically grounded categorization allowed for comprehensive representation of patient status while preserving interpretability. All categories and selected features were independently reviewed and validated by a board-certified intensivist to ensure clinical accuracy and relevance for mortality prediction in hypertensive kidney injury.

To refine this high-dimensional feature space, we implemented a two-stage selection strategy combining random forest importance ranking and mutual information (MI) filtering. These methods were chosen for their complementary strengths: the former captures non-linear relationships and interaction effects using an embedded model framework, while the latter quantifies the direct dependency between each feature and the target outcome without relying on model assumptions.

In the random forest ranking stage, a random forest classifier was trained on the full set of candidate features, and Gini importance scores were computed for each variable. The importance of a feature $x_i$ was calculated based on the average decrease in Gini impurity across all decision trees in the forest:

\begin{equation}
I(x_i) = \sum_{t \in T} \sum_{n \in \text{nodes}(t)} \mathbb{1}_{\{ \text{split}(n) = x_i \}} \cdot p(n) \cdot \Delta G(n)
\end{equation}

where $T$ is the set of all trees in the forest, $n$ denotes a decision node, $\text{split}(n)$ is the feature used at node $n$, $p(n)$ is the proportion of training samples reaching node $n$, and $\Delta G(n)$ is the decrease in Gini impurity at node $n$. Features that consistently contributed little to impurity reduction, such as \textit{hematocrit}, \textit{FiO\textsubscript{2}}, and \textit{calcium}, were removed due to their low $I(x_i)$ values and redundancy with more stable predictors.

In the mutual information (MI) stage, we computed the MI score between each remaining feature $x_i$ and the binary outcome variable $Y$ (30-day mortality). The MI quantifies the reduction in uncertainty about $Y$ given knowledge of $x_i$:

\begin{equation}
\text{MI}(x_i, Y) = \sum_{x_i \in \mathcal{X}} \sum_{y \in \mathcal{Y}} p(x_i, y) \log \left( \frac{p(x_i, y)}{p(x_i) p(y)} \right)
\end{equation}

Features with $\text{MI}(x_i, Y)$ values near zero—indicating weak dependency—were excluded. For example, \textit{temperature}, \textit{chloride}, and \textit{hematocrit} were eliminated due to their minimal contribution to reducing outcome uncertainty.

After combining these two selection stages, we retained a compact yet clinically meaningful set of 18 features, which were again reviewed by domain experts to confirm their interpretability and relevance. These features span multiple clinical dimensions, capturing key aspects of patient physiology, organ function, comorbid conditions, therapeutic interventions, and care delivery processes. The final feature set was designed to balance statistical importance with clinical applicability, ensuring that downstream modeling remains both robust and interpretable. All selected variables were independently validated by critical care physician to ensure their clinical appropriateness and alignment with real-world ICU decision-making. The final selected features are summarized in Table~\ref{tab:final_features}.

\begin{table}[H]
\centering
\caption{\textbf{Final 18 features used for mortality prediction in ICU patients with hypertensive kidney injury}}
\label{tab:final_features}
\renewcommand{\arraystretch}{1.2}
\small
\begin{tabular}{p{4cm}|p{9.5cm}}
\hline
\rowcolor[HTML]{f7e1d7}
\textbf{Category} & \textbf{Selected Features} \\
\hline
chartevents & 
Richmond-RAS Scale, Non-Invasive Blood Pressure (systolic), Anion gap, Phosphorous, Respiratory Rate, GCS - Motor Response, Heart Rate Alarm - Low \\
\hline
procedureevents & 
Multi Lumen \\
\hline
labevents & 
Lactate, PTT, Bicarbonate, pO$_2$, apsiii \\
\hline
drug & 
Norepinephrine, Oxycodone (Immediate Release) \\
\hline
comorbidity & 
Cerebral edema, Severe sepsis with septic shock \\
\hline
admission & 
ED duration \\
\hline
\end{tabular}
\end{table}

\subsection*{Modeling}

\begin{figure}[h]
    \centering
    \includegraphics[width=1\linewidth]{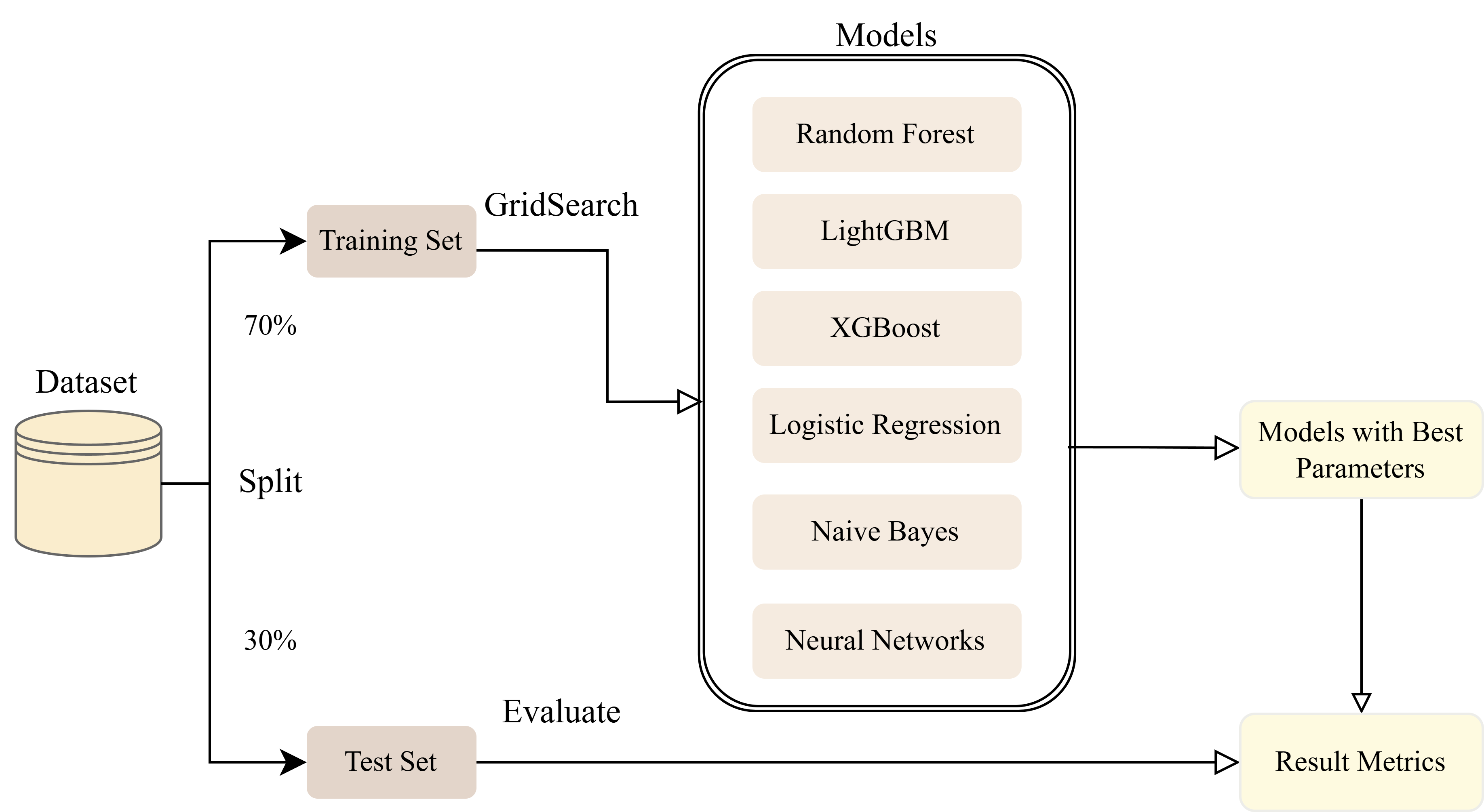}
    \caption{\textbf{Machine learning models implemented for predicting 30-day in-hospital mortality among ICU patients with hypertensive kidney disease.}}

    \label{fig:model}
\end{figure}

To predict 30-day in-hospital mortality among patients with hypertensive kidney injury, we constructed a supervised machine learning pipeline integrating multiple classification algorithms with robust validation and evaluation procedures. The dataset was split into a training set (70\%) and a held-out test set (30\%), ensuring stratified class balance. Within the training set, hyperparameter tuning and model optimization were performed using five-fold stratified cross-validation coupled with exhaustive grid search. The overall modeling workflow is illustrated in Figure~\ref{fig:model}.

Three tree-based gradient boosting frameworks—CatBoost, LightGBM, and XGBoost—were prioritized for their demonstrated effectiveness in modeling non-linear patterns and complex feature interactions in clinical datasets\cite{gao2022prediction}. These models build ensembles of decision trees in a stage-wise fashion to minimize residual errors. CatBoost was particularly suited for our dataset due to its native handling of categorical inputs and the use of ordered boosting to prevent target leakage. Meanwhile, LightGBM and XGBoost provided fast, scalable solutions with fine-grained control over tree depth, regularization parameters (\texttt{reg\_alpha}, \texttt{reg\_lambda}), learning rate, and sampling strategies, allowing fine-tuned balance between bias and variance.

For interpretability, we incorporated logistic regression as a baseline model. Both L1 (lasso) and L2 (ridge) regularization schemes were evaluated to mitigate overfitting and multicollinearity, with optimal inverse regularization strength ($C$) determined via nested cross-validation\cite{si2025retrospective}. Although linear in nature, logistic regression enables transparent inspection of variable effects through model coefficients, offering direct clinical insight.

A probabilistic Naïve Bayes classifier was also developed to serve as a lightweight benchmark. This model assumed feature independence and modeled continuous variables using Gaussian likelihoods. While its simplicity limits its flexibility in ICU settings, it provides a reference for evaluating the added value of more complex models\cite{fan2025developmentinteractivenomogramspredicting}.

To explore the potential of neural representations, we designed a shallow feedforward neural network with one hidden layer using ReLU activation and a sigmoid output node\cite{ren2019hybrid}. Training was performed with the Adam optimizer and binary cross-entropy loss. Hyperparameters—including learning rate, dropout ratio, and batch size—were optimized to promote convergence while avoiding overfitting. Despite the expressive capacity of neural networks, their interpretability and reliance on larger datasets warrant careful application in clinical domains.

Across all models, regularization strategies were applied in accordance with each algorithm’s architecture. These included L1/L2 penalties, dropout regularization, early stopping, and subsampling techniques to reduce variance and enhance generalizability.

Model performance was evaluated using a comprehensive suite of metrics to ensure both statistical robustness and clinical applicability. The primary metric was the area under the receiver operating characteristic curve (AUROC), given its threshold independent discrimination capacity. However, because mortality prediction involves significant class imbalance, additional performance indicators were included. Accuracy provided a general sense of correctness but was insufficient in isolation. The F1-score was calculated to balance precision and recall, especially important in scenarios where both false positives and false negatives have clinical consequences. Sensitivity (recall) assessed the model’s ability to identify high-risk patients—crucial in minimizing missed critical events—while specificity evaluated its precision in ruling out mortality risk, thus preventing overtreatment. Positive predictive value (PPV) and negative predictive value (NPV) were also reported to contextualize the reliability of predictions for clinical decision-making. To account for sampling variability, 95\% confidence intervals for AUROC were computed using 2{,}000 bootstrap resamples. This multifaceted evaluation framework ensured that model selection was guided not only by statistical performance but also by clinical relevance in high-stakes ICU settings.

\subsection*{Statistical analyses}

To ensure the statistical validity, clinical interpretability, and decision-level reliability of our model, we implemented a comprehensive suite of analyses targeting five core objectives: (1) to verify the baseline equivalence of training and testing cohorts using t-tests; (2) to quantify individual feature contributions through ablation analysis; (3) to characterize nonlinear effects via Accumulated Local Effects; (4) to enable individualized prediction interpretation through SHAP; and (5) to estimate predictive uncertainty using posterior distribution sampling.

We first assessed the statistical equivalence of the training and testing sets by conducting two-sided independent-sample t-tests across major clinical variables, including ED duration, anion gap. Welch's correction was applied when variance homogeneity could not be assumed. The t-statistic was computed as:

\begin{equation}
t = \frac{\bar{x}_1 - \bar{x}_2}{\sqrt{\frac{s_1^2}{n_1} + \frac{s_2^2}{n_2}}},
\end{equation}

This analysis ensured that the stratified sampling procedure did not introduce distributional imbalances, thereby reinforcing the validity of performance comparisons and reducing sampling-induced bias.

To evaluate the marginal predictive utility of individual features, an ablation analysis was performed by systematically removing each variable from the final model and re-estimating the area under the  AUROC. This process quantified the contribution of each feature to overall performance. Notably, organ function–related features such as \textit{pO\textsubscript{2}} and drug administration markers (e.g. Norepinephrine) demonstrated high performance sensitivity upon exclusion, consistent with their established roles in acute pathophysiology.

To capture nonlinear and non-additive effects of continuous features, we employed ALE, a method designed to evaluate feature influence while mitigating issues of feature collinearity\cite{li2020cellular}. The ALE function for a given feature \(x_j\) is expressed as:

\begin{equation}
ALE_j(z) = \int_{z_0}^{z} \mathbb{E}_{X_{\setminus j}} \left[ \frac{\partial f(X)}{\partial x_j} \Big| x_j = s \right] ds,
\end{equation}

ALE visualizations allowed us to identify clinically meaningful risk patterns, including threshold effects and non-monotonic relationships, such as the U-shaped mortality curves frequently observed in respiratory and renal parameters. These patterns are valuable for setting physiologic thresholds in ICU triage and intervention protocols.

For patient-specific explainability, SHAP were applied to decompose model predictions into additive feature contributions\cite{si2025machine}. SHAP values, grounded in cooperative game theory, are defined as:

\begin{equation}
\phi_i = \sum_{S \subseteq F \setminus \{x_i\}} \frac{|S|! (|F|-|S|-1)!}{|F|!} \left[f(S \cup \{x_i\}) - f(S)\right],
\end{equation}

This allowed us to trace each individual risk prediction back to contributing features, enabling granular clinical auditability and enhancing end-user trust. SHAP plots also provided global feature rankings, corroborating biological plausibility and aiding in clinical prioritization.

To account for uncertainty in individual-level risk predictions, we integrated a Bayesian posterior estimation framework using the DiffeRential Evolution Adaptive Metropolis algorithm. This method samples from the posterior distribution over model parameters to generate full predictive intervals\cite{li2020prediction}. The posterior predictive distribution was computed as:

\begin{equation}
p(y \mid X) = \int p(y \mid \theta, X) \cdot p(\theta \mid X) \, d\theta,
\end{equation}

This probabilistic approach enabled us to move beyond point estimates by quantifying uncertainty around each mortality prediction—an essential capability for ICU clinicians faced with ambiguous or borderline-risk cases. By surfacing predictive ambiguity, this framework supports more cautious and personalized care decisions.

Collectively, this ensemble of statistical analyses—encompassing population-level validation, feature-level contribution, local interpretability, and uncertainty quantification—ensured that the developed machine learning framework is not only accurate but also robust, interpretable, and clinically actionable in the context of HKD patient care.

\section*{Results}
\subsection*{Cohort Characteristics and Statistical Comparison}

The dataset analyzed in this study comprises critically ill patients diagnosed with hypertensive kidney disease (HKD) and admitted to the ICU. To facilitate model development and evaluation, the cohort was randomly divided using stratified sampling into a training set (70\%, \(n = 956\)) and a test set (30\%, \(n = 410\)), ensuring that class distributions remained consistent across partitions. In addition, patients were grouped based on 30-day mortality outcomes to compare baseline characteristics between survivors and non-survivors.

Table~\ref{tab:cohort_comparison_results} summarizes the descriptive statistics and p-values for 18 key clinical features across the training and test cohorts. No statistically significant differences were found between groups (\(p > 0.05\) for all features), indicating strong consistency in feature distribution. This suggests that the training and testing datasets are well-balanced and representative, which strengthens the model's external validity and supports generalization to unseen ICU HKD cases.

In contrast, Table~\ref{tab:cohort_comparison_results_1} presents comparisons between survivors and non-survivors, revealing multiple statistically significant differences. For instance, survivors had lower lactate levels (1.96 vs. 2.84), reduced anion gap (14.90 vs. 17.10), shorter PTT, and higher GCS motor response scores compared to non-survivors. Hemodynamic indicators such as higher systolic blood pressure and lower respiratory rate were more favorable in the survival group. Moreover, survivors were less likely to require vasopressors (e.g., norepinephrine), exhibited better oxygenation (higher pO\textsubscript{2}), and demonstrated more normal mental status (higher Richmond-RAS scores).

These findings reinforce the biological plausibility and clinical relevance of the selected features. Many of the statistically significant predictors correspond to established mechanisms of poor prognosis in HKD patients—namely, metabolic acidosis, neurologic compromise, respiratory distress, and circulatory shock. Their prominence in the model underscores its alignment with pathophysiological insights and highlights its interpretability.

Moreover, the similarity between the training and test sets confirms the robustness of the model training process. By using well-matched partitions and validating on independent data, we reduced the risk of sampling bias or overfitting. The mortality-based comparison also enhances transparency and interpretability, offering clinicians a direct connection between input features and real-world outcomes in HKD patients. Together, these strategies contribute to a trustworthy and clinically actionable prediction framework for ICU mortality risk stratification.

\begin{table}[H]
\caption{\textbf{T-test Comparison of Feature Distributions between Training and Test Sets.}}
\label{tab:cohort_comparison_results}
\small
\renewcommand{\arraystretch}{1.2}
\rowcolors{2}{white}{white}
\begin{tabularx}{\textwidth}{l|l|X|X|l}
\hline
\rowcolor[HTML]{f7e1d7}
\textbf{Feature} & \textbf{Unit} & \textbf{Training Set} & \textbf{Test Set} & \textbf{P-value} \\ \hline
Richmond-RAS Scale & -- & -1.02 (1.29) & -0.98 (1.27) & 0.549 \\ \hline
Lactate & mmol/L & 2.08 (1.16) & 2.03 (1.06) & 0.483 \\ \hline
PTT & sec & 39.07 (17.19) & 38.13 (16.25) & 0.333 \\ \hline
GCS - Motor Response & score & 5.13 (1.21) & 5.17 (1.16) & 0.577 \\ \hline
Anion gap & mmol/L & 15.18 (4.08) & 14.95 (4.08) & 0.324 \\ \hline
Phosphorous & mg/dL & 4.19 (1.41) & 4.08 (1.29) & 0.191 \\ \hline
pO2 & mmHg & 129.54 (66.28) & 132.18 (64.12) & 0.490 \\ \hline
Respiratory Rate & breaths/min & 18.95 (3.59) & 18.97 (3.24) & 0.928 \\ \hline
Bicarbonate & mmol/L & 21.66 (3.38) & 21.90 (3.65) & 0.253 \\ \hline
APSIII & score & 51.12 (19.29) & 50.50 (18.02) & 0.566 \\ \hline
Non-Invasive BP Systolic & mmHg & 123.27 (19.05) & 122.92 (18.83) & 0.758 \\ \hline
ED duration & hours & 4.09 (6.77) & 4.73 (5.60) & 0.071 \\ \hline
HR Alarm - Low & bpm & 50.93 (5.26) & 50.61 (4.73) & 0.264 \\ \hline
Norepinephrine & binary & 0.38 (0.49) & 0.37 (0.48) & 0.627 \\ \hline
Oxycodone (IR) & binary & 0.46 (0.50) & 0.46 (0.50) & 0.869 \\ \hline
Severe Sepsis with Shock & binary & 0.17 (0.38) & 0.16 (0.36) & 0.420 \\ \hline
Cerebral Edema & binary & 0.05 (0.22) & 0.05 (0.22) & 0.934 \\ \hline
Multi Lumen & binary & 0.34 (0.47) & 0.34 (0.48) & 0.977 \\ \hline
\end{tabularx}
\begin{flushleft}
\textit{Note}: This table summarizes statistical comparisons between the training and test cohorts. Continuous variables are expressed as mean (standard deviation). P-values are derived from two-sided t-tests, with significance set at $p<0.05$.
\end{flushleft}
\end{table}

\begin{table}[H]
\caption{\textbf{T-test Comparison of Feature Distributions Between Non-Survival and Survival.}}
\label{tab:cohort_comparison_results_1}
\small
\renewcommand{\arraystretch}{1.2}
\rowcolors{2}{white}{white}
\begin{tabularx}{\textwidth}{l|l|X|X|l}
\hline
\rowcolor[HTML]{f7e1d7}
\textbf{Feature} & \textbf{Unit} & \textbf{Survival} & \textbf{Non-Survival} & \textbf{P-value} \\ \hline
Richmond-RAS Scale & -- & -0.87 (1.13) & -2.03 (1.80) & < 0.001 \\ \hline
Lactate & mmol/L & 1.96 (0.88) & 2.84 (2.17) & < 0.001 \\ \hline
PTT & sec & 38.28 (16.34) & 44.46 (21.48) & 0.003 \\ \hline
GCS - Motor Response & score & 5.26 (1.06) & 4.25 (1.69) & < 0.001 \\ \hline
Anion gap & mmol/L & 14.90 (3.86) & 17.10 (4.98) & < 0.001 \\ \hline
Phosphorous & mg/dL & 4.10 (1.33) & 4.76 (1.78) & < 0.001 \\ \hline
pO2 & mmHg & 132.62 (67.58) & 108.46 (52.13) & < 0.001 \\ \hline
Respiratory Rate & breaths/min & 18.74 (3.48) & 20.41 (3.94) & < 0.001 \\ \hline
Bicarbonate & mmol/L & 21.86 (3.29) & 20.24 (3.61) & < 0.001 \\ \hline
APSIII & score & 49.36 (18.07) & 63.19 (22.82) & < 0.001 \\ \hline
Non-Invasive BP Systolic & mmHg & 124.29 (19.28) & 116.29 (15.80) & < 0.001 \\ \hline
ED duration & hours & 4.02 (6.91) & 4.54 (5.76) & 0.372 \\ \hline
HR Alarm - Low & bpm & 51.20 (5.34) & 49.11 (4.26) & < 0.001 \\ \hline
Norepinephrine & binary & 0.33 (0.47) & 0.70 (0.46) & < 0.001 \\ \hline
Oxycodone (IR) & binary & 0.50 (0.50) & 0.24 (0.43) & < 0.001 \\ \hline
Severe Sepsis with Shock & binary & 0.14 (0.35) & 0.41 (0.49) & < 0.001 \\ \hline
Cerebral Edema & binary & 0.04 (0.20) & 0.11 (0.32) & 0.018 \\ \hline
Multi Lumen & binary & 0.32 (0.47) & 0.48 (0.50) & 0.002 \\ \hline
\end{tabularx}
\begin{flushleft}
\textit{Note}: This table compares patients with and without 30-day survival. Continuous features are shown as mean (standard deviation). P-values were calculated using two-sided t-tests with a significance threshold of $p<0.05$.
\end{flushleft}
\end{table}

\subsection*{Feature Contribution Analysis}

\begin{figure}[H]
    \centering
    \includegraphics[width=0.95\linewidth]{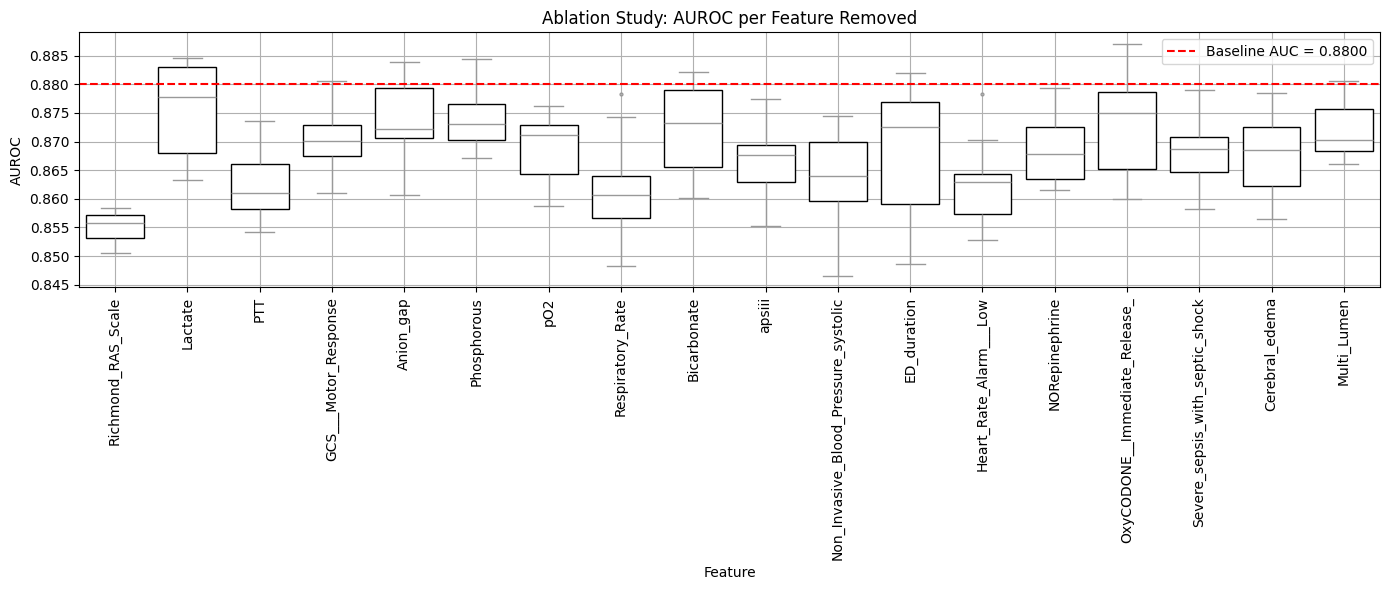}
    \caption{\textbf{Impact of Feature Removal on LR Model Performance.}}
    \label{fig:ablation_analysis}
\end{figure}

To evaluate individual feature contributions for predicting vancomycin-associated creatinine elevation in critically ill patients, we conducted an ablation analysis. As illustrated in Figure~\ref{fig:ablation_analysis}, each feature was systematically removed, and a logistic regression classifier was retrained using bootstrap sampling to assess its marginal impact on AUROC. The red dashed line represents the baseline AUROC (0.8800) achieved when all features were included. This analysis aimed to quantify the predictive utility of each variable independently of specific model architectures, providing information on which clinical factors most strongly contribute to the evaluation of nephrotoxicity risk.

Notably, the exclusion of Richmond-RAS Scale led to the most substantial decline in model performance (AUROC dropping to approximately 0.855), indicating that neurological status plays a dominant role in identifying patients at higher risk of creatinine elevation following vancomycin exposure. This finding aligns with clinical observations that altered consciousness may reflect systemic illness severity or compromised physiological reserve.

Other features with considerable impact included PTT and lactate, which showed notable performance decreases when removed (AUROC approximately 0.869 and 0.868 respectively), highlighting the importance of coagulation status and metabolic markers in nephrotoxicity risk assessment. In contrast, features such as anion gap, phosphorus, and several physiological monitoring variables showed minimal impact when excluded, suggesting these variables provide limited unique predictive information in this clinical context.

Overall, the ablation results demonstrate clear hierarchical importance among clinical variables, with neurological assessment, coagulation markers, and metabolic indicators providing the strongest individual contributions to vancomycin nephrotoxicity prediction. These findings support the biological plausibility of the feature set and provide valuable insights for clinical risk factor prioritization.

\subsection*{Model Performance Evaluation and Clinical Interpretability}

\begin{figure}[H]
\centering
\includegraphics[width=0.95\linewidth]{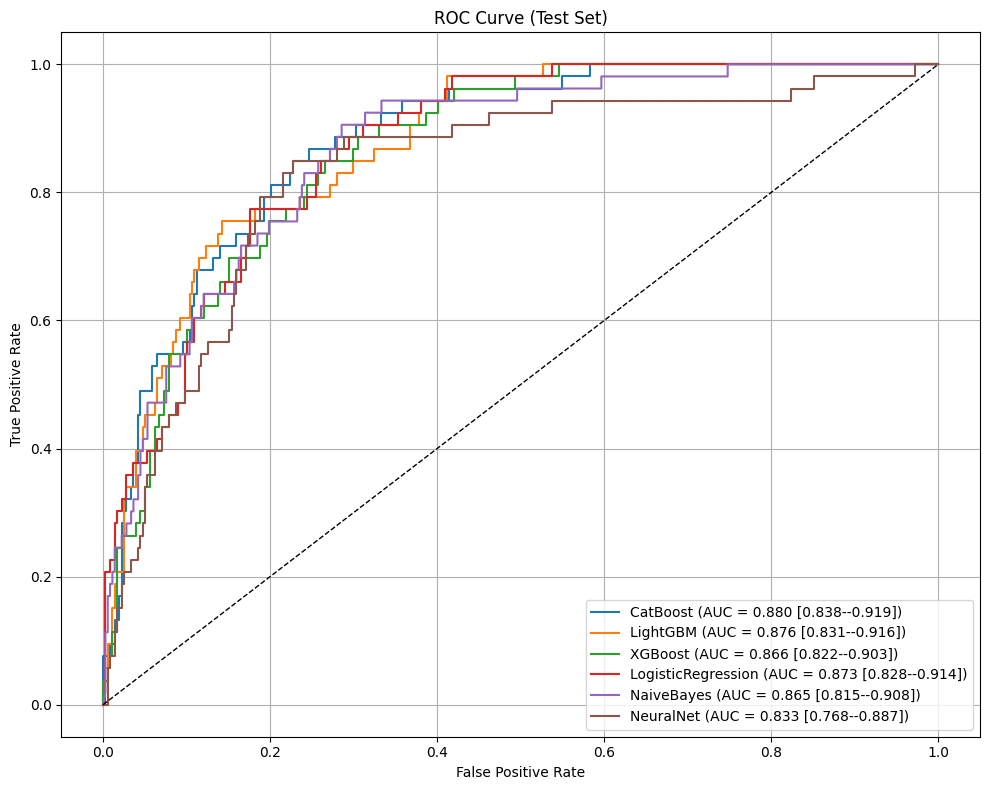}
\caption{\textbf{AUROC Curves for Model Performance in the Test Set.}}
\label{fig:roc_test}
\end{figure}

To evaluate the capacity of different algorithms to predict vancomycin-associated creatinine elevation among ICU patients, we tested six widely used machine learning models. Performance metrics on the test set—including AUROC, sensitivity, specificity, F1-score, and predictive values—are summarized in Table~\ref{tab: Results of the Test Set}. ROC curves for the test set are illustrated in Figure~\ref{fig:roc_test}.

Our dataset exhibited moderate class imbalance with 12.8\% nephrotoxicity cases (175 of 1,366 patients in the test set). To address this challenge, we employed a sensitivity-prioritized evaluation strategy rather than standard probability thresholds. CatBoost achieved balanced performance with sensitivity of 0.811 and specificity of 0.798, while the positive predictive value (PPV) of 0.374 represents nearly three-fold improvement over baseline prevalence. The high negative predictive value (NPV) of 0.966 provides strong confidence for low-risk predictions, demonstrating that our approach successfully handled class imbalance by prioritizing clinically relevant metrics over accuracy alone.

Rather than using fixed probability thresholds, we employed a clinically-motivated threshold selection strategy. For each model, we identified operating points where sensitivity remained at least 0.80 and exceeded specificity, prioritizing the detection of high-risk patients while maintaining reasonable precision. This approach reflects clinical priorities in nephrotoxicity prediction, where missing a patient at risk for kidney injury (false negative) carries greater clinical consequences than false alarms that prompt additional monitoring.

Among the models evaluated, CatBoost achieved the highest AUROC of 0.880 (95\% CI: 0.838--0.919), indicating strong discriminatory ability. Under our sensitivity-prioritized threshold selection, CatBoost maintained excellent performance with sensitivity of 0.811, specificity of 0.798, and an F1-score of 0.512. The model achieved an accuracy of 0.800, demonstrating robust overall classification performance despite the class imbalance challenge.

Critically, CatBoost's performance demonstrates genuine predictive skill beyond baseline rates. The positive predictive value (PPV) of 0.374 represents a substantial improvement over the 12.8\% baseline prevalence, indicating that patients flagged as high-risk are approximately three times more likely to develop creatinine elevation than random selection. More importantly, the negative predictive value (NPV) of 0.966 substantially exceeds what a naive classifier would achieve, providing high confidence that patients predicted as low-risk will not develop nephrotoxicity.

From a clinical perspective, this level of performance is particularly valuable in ICU settings where vancomycin is commonly used but carries well-documented nephrotoxic potential. The high sensitivity ensures that 81\% of patients who will develop creatinine elevation are identified prospectively, enabling timely interventions such as enhanced monitoring, dose adjustments, or consideration of alternative therapies. Simultaneously, the strong specificity (79.8\%) helps avoid unnecessary interventions in low-risk patients.

CatBoost was selected as the final model not only for its superior test performance but also for its robustness to missing data and ability to handle heterogeneous ICU feature sets. Its gradient boosting framework with ordered boosting strategy reduces overfitting risks while maintaining interpretability through native support for categorical variables. The model's reliance on physiologically meaningful predictors enhances clinical trust and facilitates integration into existing clinical workflows.

In summary, the sensitivity-prioritized evaluation strategy and systematic handling of class imbalance resulted in a clinically relevant prediction framework. The model provides actionable risk stratification that can support early detection of vancomycin-associated nephrotoxicity while maintaining practical utility in real-world ICU settings.

\begin{figure}[H]
\centering
\includegraphics[width=0.95\linewidth]{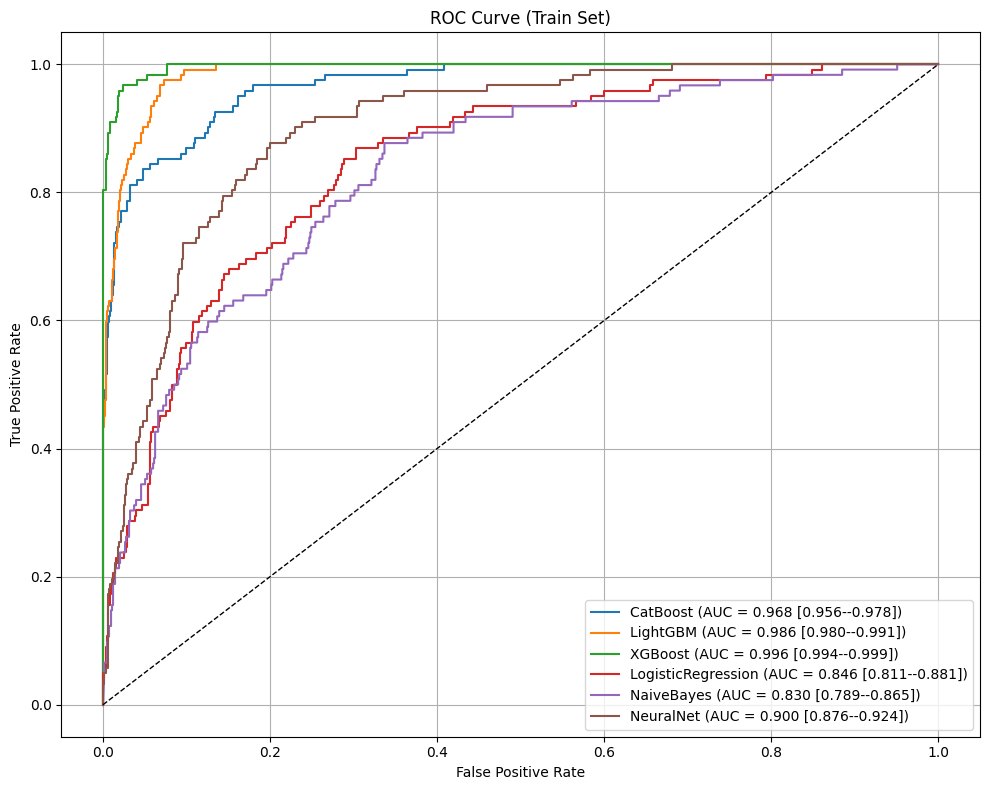}
\caption{\textbf{AUROC Curves for Model Performance in the Training Set.}}
\label{fig:roc_train}
\end{figure}

\begin{table}[H]
\renewcommand{\arraystretch}{1.2}
\centering
\caption{\textbf{Performance Comparison of Different Models in the Training Set.}}
\resizebox{\textwidth}{!}{
\begin{tabular}{l|l|l|l|l|l|l|l}
\hline
\rowcolor[HTML]{f7e1d7}
\textbf{Model} & \textbf{AUC (95\% CI)} & \textbf{Accuracy} & \textbf{F1-score} & \textbf{Sensitivity} & \textbf{Specificity} & \textbf{PPV} & \textbf{NPV} \\ \hline
\rowcolor[HTML]{a8dadc}
CatBoost & 0.968 (0.956--0.978) & 0.952 & 0.803 & 0.770 & 0.978 & 0.839 & 0.967 \\
LightGBM & 0.986 (0.980--0.991) & 0.955 & 0.817 & 0.787 & 0.980 & 0.850 & 0.969 \\
XGBoost & 0.996 (0.994--0.999) & 0.979 & 0.917 & 0.910 & 0.989 & 0.925 & 0.987 \\
LogisticRegression & 0.846 (0.811--0.881) & 0.774 & 0.449 & 0.721 & 0.782 & 0.326 & 0.950 \\
NaiveBayes & 0.830 (0.789--0.865) & 0.783 & 0.430 & 0.639 & 0.805 & 0.324 & 0.938 \\
NeuralNet & 0.900 (0.876--0.924) & 0.788 & 0.518 & 0.893 & 0.772 & 0.365 & 0.980 \\ \hline
\end{tabular}
}
\label{tab: Results of the Training Set}
\end{table}

\begin{table}[H]
\renewcommand{\arraystretch}{1.2}
\centering
\caption{\textbf{Performance Comparison of Different Models in the Test Set.}}
\resizebox{\textwidth}{!}{
\begin{tabular}{l|l|l|l|l|l|l|l}
\hline
\rowcolor[HTML]{f7e1d7}
\textbf{Model} & \textbf{AUC (95\% CI)} & \textbf{Accuracy} & \textbf{F1-score} & \textbf{Sensitivity} & \textbf{Specificity} & \textbf{PPV} & \textbf{NPV} \\ \hline
\rowcolor[HTML]{a8dadc}
CatBoost & 0.880 (0.838--0.919) & 0.800 & 0.512 & 0.811 & 0.798 & 0.374 & 0.966 \\
LightGBM & 0.876 (0.831--0.916) & 0.739 & 0.446 & 0.811 & 0.728 & 0.307 & 0.963 \\
XGBoost & 0.866 (0.822--0.903) & 0.763 & 0.470 & 0.811 & 0.756 & 0.331 & 0.964 \\
LogisticRegression & 0.873 (0.828--0.914) & 0.756 & 0.468 & 0.830 & 0.745 & 0.326 & 0.967 \\
NaiveBayes & 0.865 (0.815--0.908) & 0.768 & 0.475 & 0.811 & 0.762 & 0.336 & 0.965 \\
NeuralNet & 0.833 (0.768--0.887) & 0.790 & 0.506 & 0.830 & 0.784 & 0.364 & 0.969 \\ \hline
\end{tabular}
}
\label{tab: Results of the Test Set}
\end{table}

\subsection*{Posterior Distribution and Uncertainty-Aware Mortality Risk Prediction}

To move beyond deterministic prediction and better reflect real-world uncertainty, we applied the DREAM algorithm to derive posterior distributions of 30-day mortality risk for ICU patients with HKD. Unlike point estimates, this approach generates a full probability distribution, enabling clinicians to visualize and interpret not only the most likely outcome, but also the range and confidence of that estimate - a critical feature in high-stakes ICU decision making.

The input priors were constructed from the empirical means and standard deviations of key features in the non-survivor subgroup (Table~\ref{tab:cohort_comparison_results_1}). These included elevated APS III scores, increased respiratory support, and reduced GCS motor responses, all of which align with clinically recognized deterioration patterns in HKD patients. This design simulates a prototypical high-risk ICU presentation based on actual patient data, rather than artificial perturbation or averaged profiles.

\begin{figure}[H]
    \centering
    \includegraphics[width=0.85\linewidth]{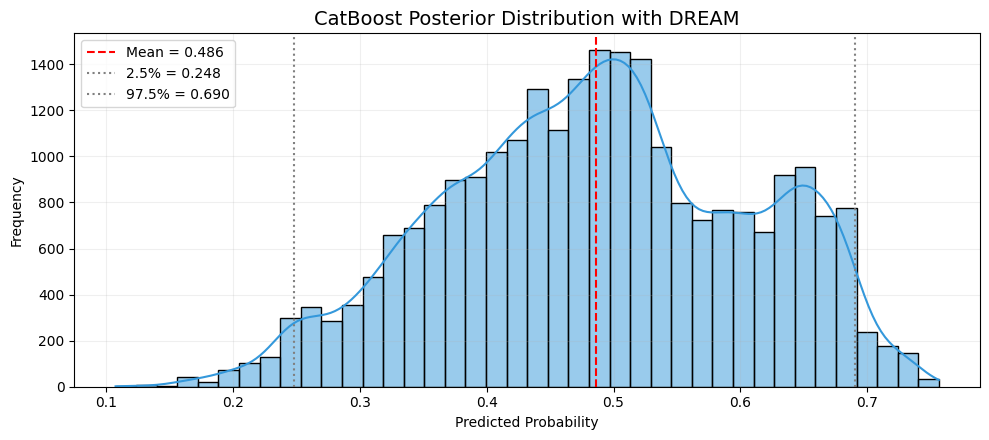}
    \caption{\textbf{Posterior distribution of 30-day mortality for a high-risk HKD ICU patient.}}
    \label{fig:uq_catboost}
\end{figure}

As shown in Figure~\ref{fig:uq_catboost}, the posterior distribution is skewed to the right, with a mean predicted mortality of 0.486 and a credible interval 95\% ranging from 0.248 to 0.690. In a practical ICU context, such a distribution offers considerable interpretive advantage. For example, a clinician who evaluates a patient with frail vasopressor-dependent HKD can recognize that even if the mean risk is less than 0.5, the upper bound exceeds 65\%, indicating substantial uncertainty and justifying precautionary actions such as early palliative discussions or more intensive monitoring. 

This is particularly valuable in cases of borderline physiology, where a simple binary 'high-risk' label can obscure the underlying distribution. For example, two patients with the same mean predicted risk may differ significantly in their credible intervals: one narrow and confident, another wide and uncertain, which implies very different clinical management paths.

Importantly, DREAM does not require model retraining but leverages the trained CatBoost model and importance sampling to generate the posterior. This computational efficiency facilitates real-time deployment in clinical workflows. By conditioning predictions on prior knowledge from non-survivors, the system not only replicates expert intuition but also provides a quantifiable confidence framework for nuanced triage decisions.

In sum, posterior-based inference with DREAM offers a statistically principled and clinically grounded mechanism for individualized risk communication in patients with HKD, advancing interpretability and trust in ML-driven ICU decision support.

\subsection*{ALE Analysis and Clinical Interpretability}

\begin{figure}[H]
    \centering
    \includegraphics[width=0.85\linewidth]{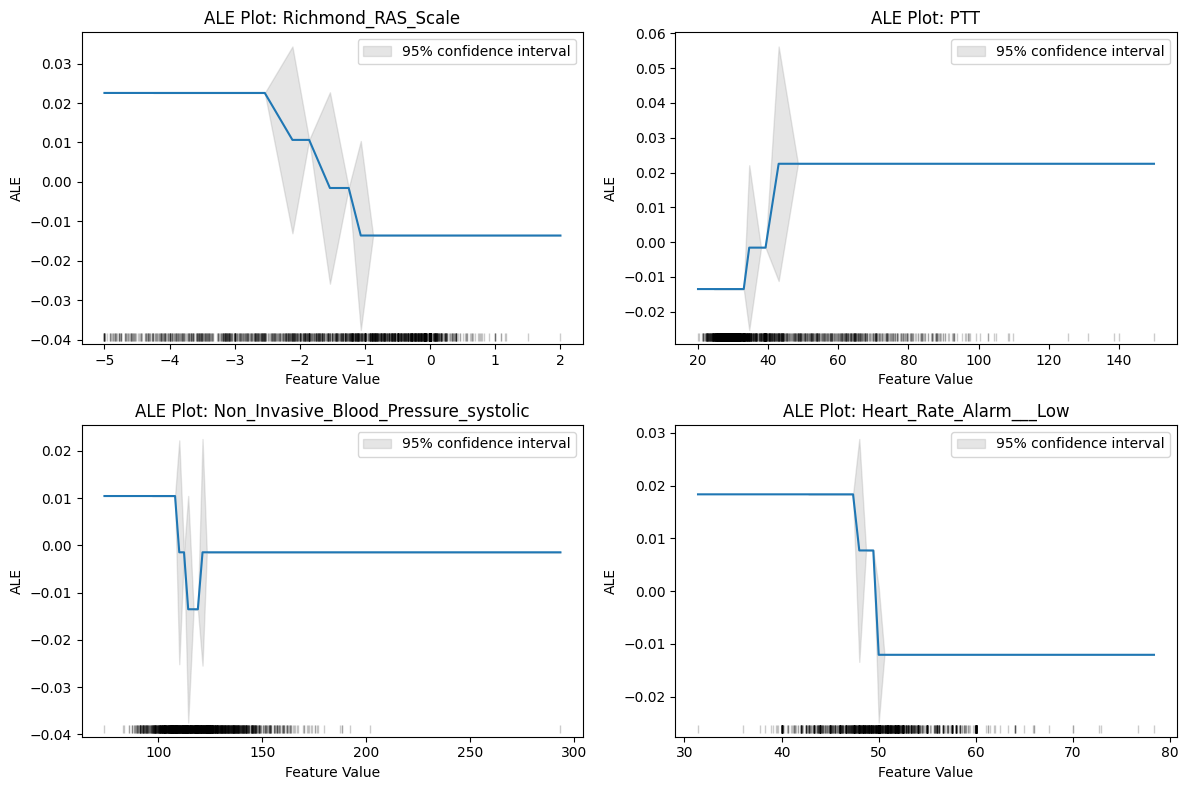}
    \caption{\textbf{ALE plots for top features in HKD ICU patients.}}
    \label{fig:ale_analysis}
\end{figure}

To further evaluate the clinical plausibility of the CatBoost model in predicting 30-day mortality among ICU patients with HKD, we performed an accumulation of local effects (ALE) analysis on four statistically significant and medically relevant variables: Richmond-RAS Scale, PTT, Non-Invasive Systolic Blood Pressure, and Heart Rate Alarm – Low. The plots in Figure~\ref{fig:ale_analysis} illustrate how changes in these individual features affect model predictions, adjusted for the conditional distribution of the remaining covariates.

The ALE plot for the Richmond-RAS Scale shows a gradual downward slope as values move from -1 to -5, suggesting that deeper levels of sedation or impaired consciousness are associated with higher predicted mortality. This trend aligns with the clinical understanding that HKD patients who are non-responsive or deeply sedated often have severe metabolic encephalopathy or are under heavy sedation due to ventilatory distress. The left-shifted ALE effect at RASS $\leq -3$ reinforces its predictive relevance.

For PTT, the curve rises sharply between 30 and 50 seconds, before plateauing beyond 60 seconds. This suggests that mild to moderate coagulopathy significantly increases mortality risk, consistent with disseminated intravascular coagulation (DIC) or hepatic dysfunction commonly seen in HKD-related critical illness. The confidence interval is widest in this range, reflecting limited support from extreme values, yet the effect direction remains biologically plausible.

Systolic blood pressure, captured by the non-invasive measurement, reveals a sharp negative ALE effect when values fall below 110 mmHg. This suggests that hypotension—even modest—strongly contributes to increased mortality risk, which matches the physiological vulnerability of HKD patients with impaired autoregulation and poor cardiovascular reserve. In practical terms, even transient blood pressure drops may necessitate early vasopressor intervention.

The ALE curve for Heart Rate Alarm – Low exhibits a modest negative slope starting at 50 bpm, indicating that bradycardia below this threshold is associated with increased predicted mortality. Clinically, this may reflect conduction abnormalities or autonomic instability common in advanced renal disease. The clustering of values around 50–55 bpm also indicates strong model support in this region.

Overall, these ALE results support both the biological credibility and local interpretability of the model. Smooth and directional trends confirm that the model risk estimates respond to known markers of deterioration in the ICU in patients with HKD. Moreover, the effects are grounded in well-distributed feature ranges, enhancing confidence that these insights are not driven by outliers or spurious correlations.

\subsection*{SHAP-Based Feature Attribution and Clinical Contextualization}

To enhance interpretability of the CatBoost model and bridge statistical inference with bedside relevance, we employed SHAP analysis to quantify the marginal impact of each feature on the model's prediction of 30-day mortality in ICU patients with HKD. SHAP values represent the contribution of each variable to the deviation from the model’s baseline prediction, allowing a local, patient-specific interpretation grounded in cooperative game theory.

\begin{figure}[H]
    \centering
    \includegraphics[width=0.95\linewidth]{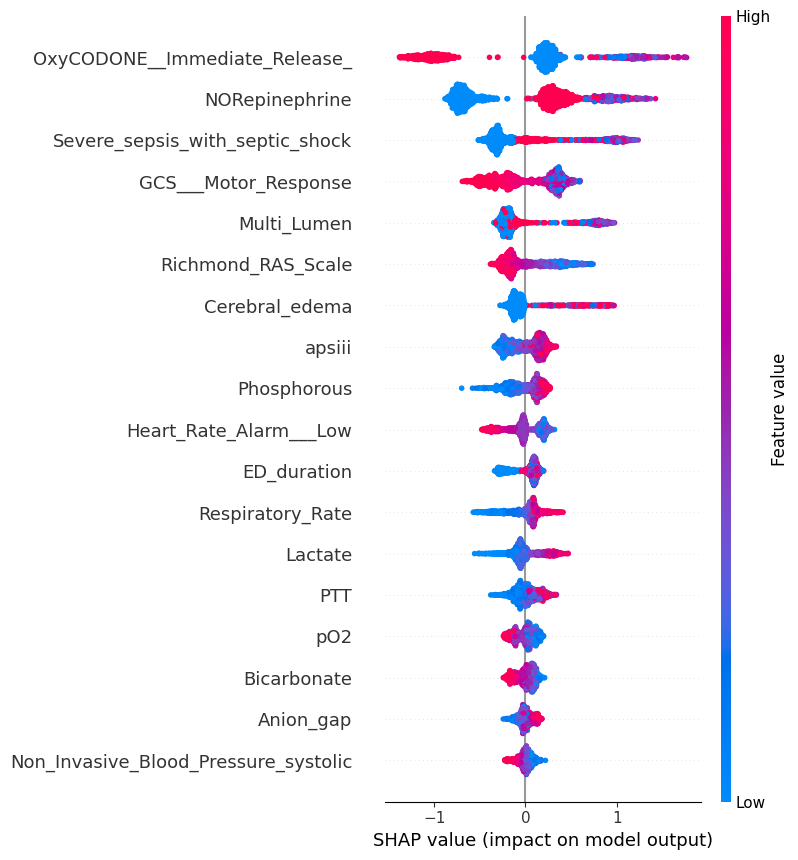}
    \caption{\textbf{SHAP summary plot showing feature contributions to predicted 30-day mortality in HKD patients.}}
    \label{fig:shap_summary}
\end{figure}

As shown in Figure~\ref{fig:shap_summary}, variables such as Oxycodone (Immediate Release), norepinephrine administration, and septic shock status demonstrated the largest absolute SHAP values, suggesting that these indicators heavily influence mortality prediction. Clinically, this is consistent with real-world ICU trajectories: the need for strong opioids often signals severe discomfort or terminal agitation, while vasopressor use and septic physiology denote hemodynamic instability—both high-risk conditions in HKD patients with limited compensatory reserve.

Neurological status (GCS motor response, Richmond-RAS) and acute complications such as cerebral edema were also prominent contributors, highlighting the model’s attention to altered consciousness—a common precursor to multiorgan dysfunction. Interestingly, physiologic markers like lactate, bicarbonate, and PTT displayed moderate SHAP impact, differing slightly from their ablation counterparts. This divergence underscores a key distinction: whereas ablation assesses global AUROC changes from feature removal, SHAP captures local, nonlinear interactions and collinearities embedded within tree ensembles.

Nonetheless, SHAP interpretations are not without limitations. The method assumes feature independence during marginal contribution estimation, which may be violated in complex ICU datasets with correlated clinical variables (e.g., lactate and pH, vasopressor use and hypotension). Therefore, SHAP should be viewed as complementary to ablation and ALE analysis rather than a definitive ranking.

From a practical standpoint, SHAP enables real-time interpretability in the ICU by tracing back each prediction to its key contributors. For instance, a predicted high mortality risk can be contextualized by elevated anion gap, low GCS, and recent vasopressor use, facilitating more nuanced conversations around prognosis, escalation, or palliative planning. The alignment between SHAP results and known clinical deterioration pathways further validates the model's face validity and its applicability to real-world critical care decision-making.

\section*{Discussion}
\subsection*{Summary of Existing Model Compilation}
This study proposes a machine learning-based framework to predict 30-day in-hospital mortality among ICU patients with HKD using early clinical features extracted from the MIMIC-IV database. By filtering adult ICU patients with HKD diagnoses and excluding active malignancies, we curated a clinically meaningful cohort for analysis. A total of 18 features—including vital signs, laboratory results, interventions, and severity scores—were selected based on their clinical importance and statistical relevance. Among multiple classifiers, CatBoost demonstrated the best overall performance, achieving an AUROC of 0.88 in the test set. In addition to accuracy, the study emphasized clinical applicability by incorporating interpretability tools such as SHAP, ALE plots, and ablation studies to provide transparent explanations of model decisions. We further introduced the DREAM algorithm to generate patient-specific uncertainty distributions, offering clinicians a more nuanced, probabilistic understanding of individual risk. Together, the pipeline combines performance, interpretability, and uncertainty estimation into a cohesive tool ready for potential real-world deployment in ICU decision-making.

\subsection*{Comparison with Prior Studies}
Although a number of models have been proposed for outcome prediction in CKD and ICU populations, few have explicitly focused on the intersection of hypertension and chronic kidney disease in the critical care setting. Xia et al.~\cite{xia2022nomogram} developed a nomogram based on a Cox proportional hazards model to estimate long-term survival in hypertensive CKD patients. While this model achieved a moderate C-index of 0.70, it was built using outpatient data and was designed for long-term prognosis rather than real-time ICU application. It also lacked the inclusion of ICU-specific variables and did not address dynamic physiological changes.

Kang et al.~\cite{kang2020ml} employed a random forest classifier to predict in-hospital mortality among ICU patients with dialysis-requiring acute kidney injury. Their model significantly outperformed traditional scoring systems such as APACHE II and SOFA, achieving an AUROC of 0.78. Although this study demonstrated the potential of machine learning in renal-critical care, it did not include hypertension as a contributing comorbidity and lacked uncertainty quantification or layered interpretability.

Li et al.~\cite{li2023sepsisaki} developed multiple machine learning models to predict mortality in ICU patients with sepsis-associated acute kidney injury using the MIMIC-IV database. Their XGBoost model achieved an AUROC of 0.794 and employed both SHAP and LIME for interpretability. The top predictors identified included SOFA score, respiratory rate, SAPS II, and age—factors consistent with known ICU risk markers. Despite its strengths, this model did not stratify patients based on chronic hypertension or CKD status and did not prioritize early-phase ICU data, thereby limiting its immediate applicability for proactive clinical intervention.

In contrast to these prior works, our study is the first to target the high-risk population of HKD patients in the ICU. Our model is specifically optimized for real-time risk prediction using only data from the first 24 hours of ICU admission, ensuring relevance for early clinical intervention. Additionally, we combine high predictive performance (AUROC ~0.88) with multilayered interpretability using SHAP values, ALE plots, and targeted ablation studies to validate the clinical relevance of key features. Finally, and most importantly, our study integrates the DREAM to provide individualized uncertainty estimates alongside point predictions, enabling clinicians to understand not only the mortality risk but also the associated confidence intervals. This feature addresses one of the major limitations of prior models, which often produce deterministic outputs without reflecting prediction reliability.

Taken together, our model bridges the gap between academic model development and practical bedside deployment. It extends the clinical utility of machine learning in critical care by offering timely, interpretable, and probabilistically informed predictions tailored to a high-risk population. In doing so, it advances the current state of literature from retrospective performance reporting to clinically actionable, trust-enhancing decision support in the ICU.

\subsection*{Limitations and Future Works}
Despite promising results, several limitations warrant consideration. First, the model was developed and validated using data from a single academic center, which may constrain its generalizability across healthcare systems with different patient demographics or ICU protocols. External validation on multicenter databases such as eICU or international registries will be essential to assess model robustness and recalibrate as needed. Second, while we focused on the first 24 hours of data to simulate early triage, this required summarizing dynamic time-series variables into static features, potentially omitting important temporal patterns. Future work could incorporate time-aware architectures, such as RNNs or transformers, to capture evolving risk. Lastly, although the model demonstrated interpretability and computational efficiency, its integration into clinical workflows and electronic health record systems remains untested. Prospective implementation studies, including user-centered design evaluations, will be necessary to assess its impact on clinical decision-making and patient outcomes. Addressing these areas will help translate our model into a reliable and generalizable tool for critical care management in HKD patients.

\section*{Conclusion}
This study presents a comprehensive and clinically grounded machine learning framework for predicting 30-day in-hospital mortality in ICU patients with HKD, a high-risk subgroup often overlooked in existing prognostic models. By leveraging early clinical features available within the first 24 hours of ICU admission, our CatBoost-based model demonstrated strong predictive performance (AUROC = 0.88) and incorporated robust post hoc interpretability through SHAP values, ALE plots, and ablation analyses. These tools not only enhance transparency but also facilitate clinical understanding of the most influential predictors, such as oxygenation status, sedation level, and hemodynamic markers.

Crucially, our model integrates the DREAM algorithm to provide individualized uncertainty estimates alongside point predictions, offering a more nuanced and trustworthy decision-support system. This probabilistic transparency addresses a key limitation of prior models that deliver deterministic outputs without contextualizing prediction reliability. In contrast to existing approaches—which have either targeted long-term outcomes in outpatient CKD cohorts or focused on general ICU populations with sepsis or acute kidney injury—our framework is specifically optimized for early triage and risk stratification in HKD patients under critical care.

Overall, this work advances the application of machine learning in critical care from retrospective risk modeling toward real-time, interpretable, and uncertainty-aware decision support. As a next step, external validation across diverse institutions and integration into clinical workflows will be essential to evaluate the model’s practical impact and generalizability. With further prospective testing and system-level deployment, the proposed framework holds strong potential to augment ICU clinicians’ decision-making and improve outcomes in this vulnerable patient population.

\section*{Author Summary}
Y.S. was responsible for the overall study design, methodological development, experimental execution, data interpretation, and initial manuscript drafting. S.C., J.F., and L.S. assisted in data preprocessing, model implementation, and contributed to manuscript revisions. E.P., K.A., and G.P. offered essential insights into study design and critically reviewed the findings. M.P. oversaw the research project, facilitated team coordination, and provided strategic direction throughout. All authors participated in reviewing and approving the final manuscript.

The authors gratefully acknowledge the Laboratory for Computational Physiology at the Massachusetts Institute of Technology for their role in developing and maintaining the MIMIC-IV database, which made this research possible.


%
%
%
\bibliography{referrences}

\end{document}